\documentclass[letterpaper, 10 pt, conference]{ieeeconf}  

\IEEEoverridecommandlockouts                              

\usepackage{fancyhdr,graphicx,amsmath,amssymb}
\usepackage[ruled,vlined]{algorithm2e}
\usepackage{tabu}

\usepackage{graphics} 
\usepackage{graphicx}
\usepackage{epsfig} 
\usepackage{amsmath} 
\usepackage{amssymb}  
\usepackage{cite}
\usepackage{multirow}
\usepackage{multicol}
\usepackage{caption}
\usepackage{acronym}
\usepackage{subcaption}
\usepackage{gensymb}

\let\labelindent\relax
\usepackage[shortlabels]{enumitem}

\usepackage[usenames,dvipsnames]{xcolor}

\usepackage{booktabs}

\DeclareMathAlphabet{\mbf}{OT1}{ptm}{b}{n}

\graphicspath{{figs/}}

\title{\LARGE \bf
What to Learn: Features, Image Transformations, or Both?

}

\author{Yuxuan Chen$^{1}$, Binbin Xu$^{1}$, Frederike Dümbgen$^{1}$, and Timothy D. Barfoot$^{1}$
\thanks{ All authors are with the University
of Toronto Institute for Aerospace Studies, University of Toronto,
Canada \texttt{\{sherry.chen, binbin.xu, frederike.dumbgen, tim.barfoot\}@robotics.utias.utoronto.ca}}%
}

\begin{document}

\include{pythonlisting}

\maketitle
\thispagestyle{empty}
\pagestyle{empty}


\begin{abstract}
Long-term visual localization is an essential problem in robotics and computer vision, but remains challenging due to the environmental appearance changes caused by lighting and seasons. While many existing works have attempted to solve it by directly learning invariant sparse keypoints and descriptors to match scenes, these approaches still struggle with adverse appearance changes. Recent developments in image transformations such as neural style transfer have emerged as an alternative to address such appearance gaps. In this work, we propose to combine an image transformation network and a feature-learning network to improve long-term localization performance. Given night-to-day image pairs, the image transformation network transforms the night images into day-like conditions prior to feature matching; the feature network learns to detect keypoint locations with their associated descriptor values, which can be passed to a classical pose estimator to compute the relative poses.  We conducted various experiments to examine the effectiveness of combining style transfer and feature learning and its training strategy, showing that such a combination greatly improves long-term localization performance. 
\end{abstract}



\section{INTRODUCTION}

The goal of long-term metric localization is to estimate the 6-DoF pose of a robot with respect to a visual map. However, long-term localization remains a challenge across drastic appearance change caused by illumination variations, such as day-night scenarios. Traditional point-based localization approaches find correspondences between local features extracted from images by applying hand-crafted descriptors (e.g., SIFT~\cite{SIFT}, SURF~\cite{SURF}, ORB~\cite{ORB}), then recover the full 6-DoF camera pose. However, such hand-crafted features are not robust under extreme appearance changes due to low repeatability. To address this, experience-based visual navigation methods~\cite{exp_based_long_term_loc, vtr, MEL} proposed to store intermediate experiences to achieve long-term localization. For instance, Multi-Experience Visual Teach \& Repeat~\cite{ vtr, MEL} retrieves the most relevant experiences to perform SURF feature matching during a more challenging repeat to bridge the appearance gap and localize to the initial taught path. 

Due to recent developments in the field of deep learning, many recent works do not rely on storing intermediate experiences, and use neural networks to predict relative poses~\cite{relocnet, CamNet, NN_Net} or absolute poses~\cite{PoseNet} from images for localization. Instead of directly learning poses from images, deep-learned interest-point detectors and descriptors \cite{mona_vision, self_sup_loc, D2Net, TILDE, superpoint, LIFT, LFNet, GN_Net, R2D2, DarkPoint, unsup_metric_reloc_trfm_csty} have gained popularity since they produce more accurate results than direct pose regression when combined with a classical pose estimator. Gridseth and Barfoot \cite{mona_vision} demonstrated the effectiveness of deep-learned features in the VT\&R framework under different illumination conditions. 

While many existing approaches have attempted to improve the feature detectors and descriptors, other recent works~\cite{adver_train_adver_cond,Clement_2020, Xu:etal:RAL2021, style_transfer_for_kp_matching, AAN_Feature_matching} proposed to modify the input images using a deep image transformation to reduce the appearance gap prior to image matching. Inspired by the recent neural style-transfer techniques, the goal of \cite{style_transfer_for_kp_matching, AAN_Feature_matching} is to use a style-transfer network that transforms the query images to resemble reference images to improve the performance of day-night feature matching results for long-term localization tasks. However, these methods optimize the transform network from scratch for each considered day and night image pair, which is not feasible for real-time deployment.

\begin{figure}[t]
    \centering
        \includegraphics[width=\linewidth]{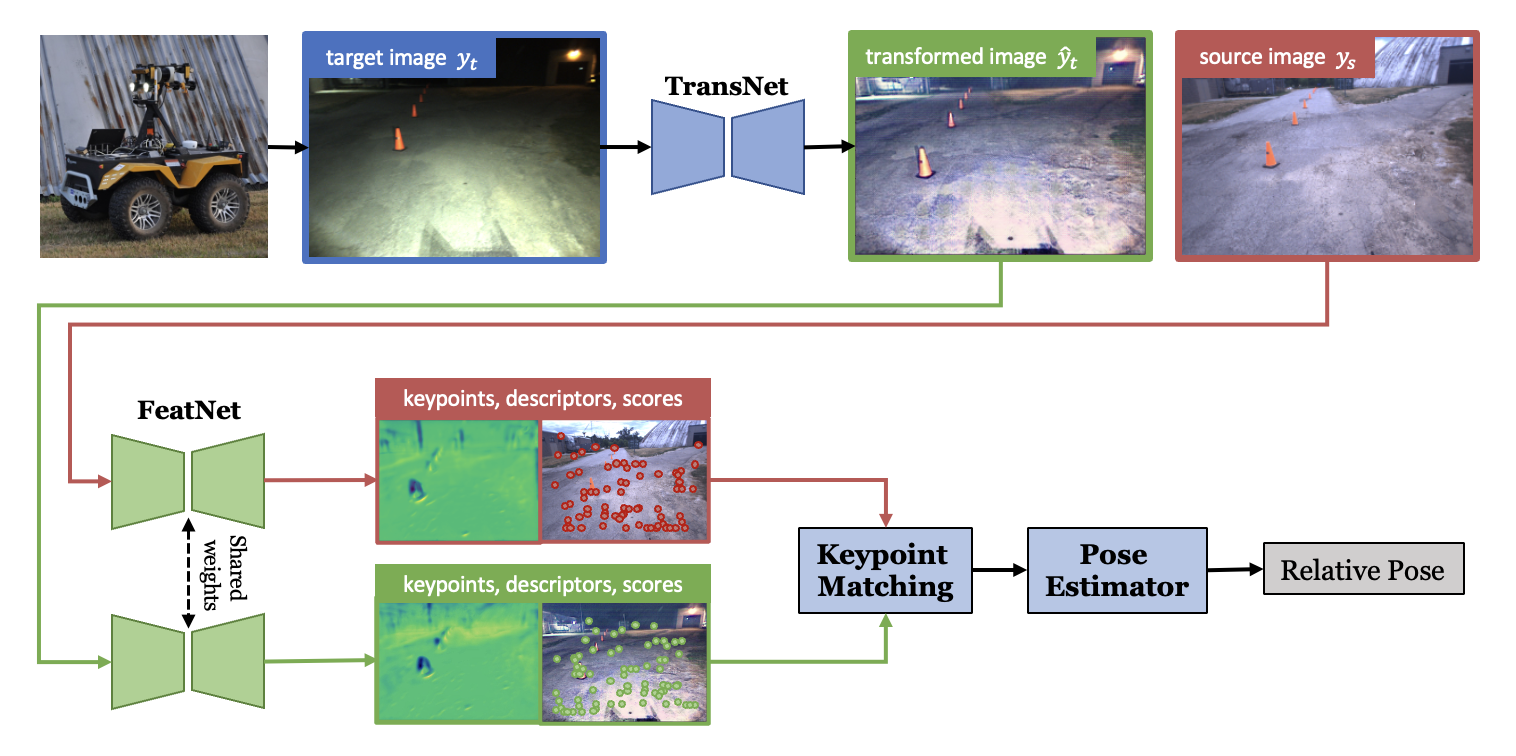}
        \caption{\textbf{High-level Overview}. TransNet transforms the target night images to day-like conditions. FeatNet predicts keypoints, descriptors, and scores on the source image and transformed target image that can be used in classical pose estimation to localize a robot despite large appearance change.}

  \label{fig:high-level overview}
\end{figure}

In this work, we attempt to answer the following question: \textit{Is it better to learn features, image transformations, or both?} We propose an end-to-end differentiable pipeline that incorporates an image transformation network and a feature learning network as shown in Figure~\ref{fig:high-level overview}. The image transformation network transforms nighttime images to resemble the style of daytime images, and the transformed images are used for feature learning. We compare our proposed method with using only the learned features or image transformation alone, and show that the combination of both components outperforms each individual part. 

    \begin{figure*}[t]
        \vspace{+0.25cm}
    	\centering
    	\includegraphics[width=0.95\linewidth]{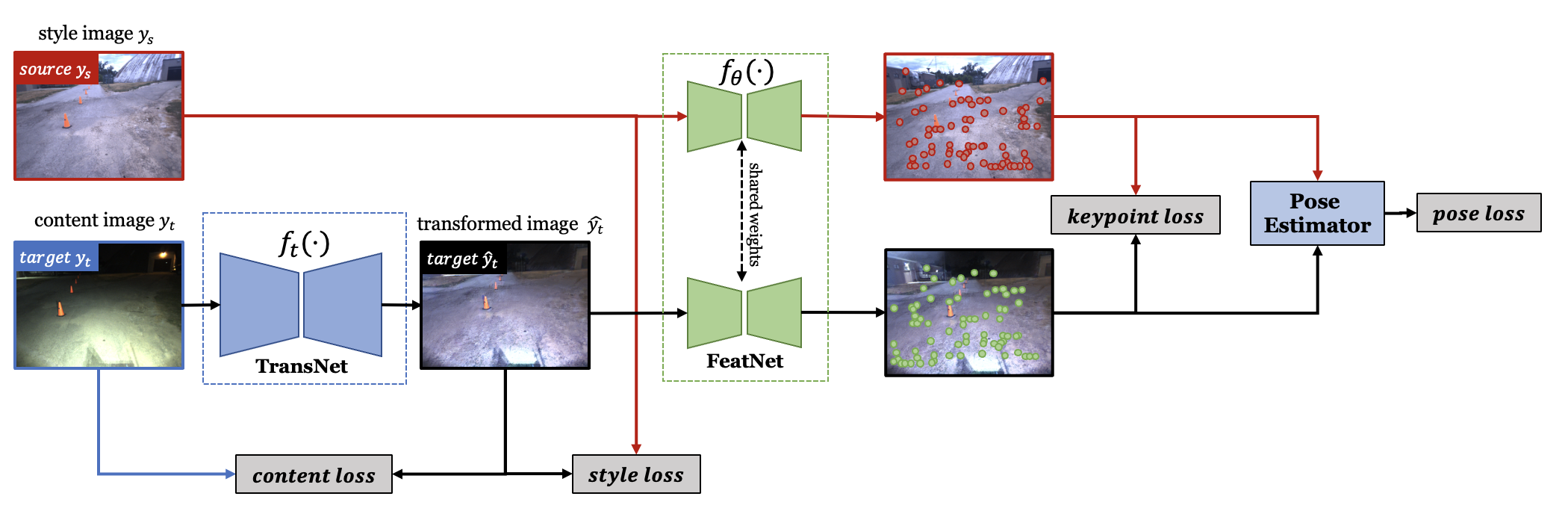}
    	\caption{\textbf{Detailed Overall Pipeline}. Given a pair of day and night images to be matched, the objective of the TransNet is to transform the night images to a day-like condition by optimizing the standard content and style loss using a fixed loss network pretrained for image classification. FeatNet takes the original style images (i.e., source images) and the transformed images (i.e., target images) as input, then outputs the keypoints location, scores, and associated descriptors for each image. The keypoints are matched in a differentiable way to generate relative poses.}
    	\label{fig:pipeline}
    \end{figure*}

\section{RELATED WORK}

Traditionally, hand-crafted features have been commonly used for visual localization~\cite{SIFT, SURF, ORB}, but suffer from low repeatability on images under dramatic appearance changes\cite{sift_surf_seasons}. Experience-based visual navigation systems\cite{exp_based_long_term_loc, work_smart_not_hard,MEL} attempt to bridge significant appearance gaps by storing multiple appearances of the same location (i.e., experiences) and choosing the most relevant experiences for feature matching during online operation.


To address the deficiencies of hand-crafted features, there have been a wide range of deep-learning-based approaches for pose estimation. Some methods tackle camera localization by training a CNN that directly learns relative poses\cite{NN_Net, relocnet, CamNet} or absolute poses\cite{PoseNet} from images. However, learning pose directly from image data can struggle with accuracy\cite{CNN_limit}. 

Other works focus on only learning the visual features or descriptors\cite{under_the_radar, mona_vision, self_sup_loc, D2Net, TILDE, superpoint, LIFT, LFNet, GN_Net, R2D2, DarkPoint}, which have successfully improved matching accuracy. The learned features can be integrated with a classical pose estimator for localization. For instance, Gridseth and Barfoot~\cite{mona_vision} and Sun et al.~\cite{DarkPoint} have proven the effectiveness of deep-learned features against illumination changes by integrating with the VT\&R pipeline for long-term visual localization. However, robustness of the learned features often come with a trade-off with respect to accuracy~\cite{D2Net}.

Instead of improving the invariance of feature detectors and descriptors under significant appearance changes, another direction is to modify the input images to reduce the appearance gap. Recent work in~\cite{Clement_2020} takes inspiration from physics-based colour constancy models and learns a nonlinear transformation from RGB to grayscale colourspace. Other approaches attempt to transform the challenging query image to resemble the reference images. Generative adversarial networks (GAN) can be useful for such image transformations. ToDayGAN~\cite{ToDayGAN} uses an unsupervised image-to-image transformation to improve the localization performance of night query images. Query images are transformed by a network to day-like conditions, where the network is trained on aligned day and night image pairs. The features are extracted from the transformed image to match against the reference image. Porav et al.~\cite{adver_train_adver_cond} proposed to use a cycle-consistency GAN with the addition of a descriptor-specific loss to help generate images that are optimized for matching without requiring aligned image pairs during pre-training. However, GAN-based methods cannot perfectly preserve the texture information that is beneficial for local-feature extraction, hence leading to limited performance gain in day-night image matching. 

Another direction is to utilize neural style transfer\cite{style_transfer} to perform image-to-image transformation. However, the traditional content and style perceptual losses used for style transfer mainly aim to reconstruct visually pleasing results, which might not result in images that are explicitly optimized for local feature extraction. Local low-level statistics of these transformed images can be drastically different compared to natural images, making it challenging to find local feature matches. To address this, \cite{style_transfer_for_kp_matching, AAN_Feature_matching} adopt style transfer and extend the traditional content and style loss with a feature-matching loss to ensure the resulting image transformation improves the matching performance. However, these methods utilize a fixed pretrained feature-detection-and-description network that is not optimized during training. In addition, these methods optimize the image transform network from scratch for each considered day and night image pair, which is not feasible for potential real-time deployment.

\section{METHODOLOGY}
\label{methodology}
    
The proposed end-to-end differentiable pipeline consists of two components: image transformation and feature learning. The image transformation network (i.e., TransNet) transforms the night images to day-like condition. The feature-learning network (i.e., FeatNet) takes the transformed day-like images as input to compute the keypoint and pose losses. The overall architecture is shown in Figure~\ref{fig:pipeline}.

\subsection{Image Transformation}
Given a pair of source and target images, the objective is to transform the target image taken under adverse conditions to be more similar to the source image in order to improve the matching performance. In this work, we specifically focus on matching night-to-day image pairs by transforming night images to day-like conditions. We adopt style transfer to solve this matching problem, where the content image corresponds to the target image (e.g., night), and the style image corresponds to the source image (e.g., day). The image transformation component consists of two networks: an image transformation network and a fixed loss network.

We denote our image transformation network to be $f_t(\cdot)$, which is composed of one encoder, five residual blocks, and one decoder similar to~\cite{style_transfer, style_transfer_for_kp_matching, AAN_Feature_matching}. The residual block is able to retain the original content information and learn additional illumination-style information of day images with shortcut connections. 

The style images are taken during daytime, denoted as $y_s$, whereas the content images, $y_t$, are taken during nighttime. The source-and-target image pair is represented by $(y_s, y_t)$. Our goal is to transform $y_t$ using $f_t(\cdot)$ to reconstruct a new image $\hat{y_t} = f_t(y_t)$, which resembles the day-like appearance style of $y_s$. 

Similar to the perceptual losses defined in~\cite{style_transfer, perceptual_style_trans}, we make use of a fixed loss network, $\phi$, to define two perceptual losses that measure the differences in content and style between images. The fixed loss network is a 16-layer VGG network~\cite{vgg16} pretrained on the ImageNet dataset~\cite{IMGNet}, where $\phi_j(y)$ denotes the $H_j \times W_j \times C_j$ feature map at layer $j$, for the input image, $y \in \mathbb{R}^{H\times W\times 3}$. For image transformation, the content image, $y_t$, is the input image and the output image, $\hat{y_t}$, should combine the content of the target image, $y_t$, with the style of the source image, $y_s$. 

\subsubsection{Content Loss}
The objective of the content loss is to enforce the similarity between the original content image, $y_t$, and the transformed image, $\hat{y_t}$, on a higher conceptual level rather than on a per pixel basis. The content loss is defined as 
\begin{equation}
    \label{eqn:content_loss}
    \mathcal{L}_{\rm content}(\hat{y_t}, y_t) = \frac{1}{H_j W_j C_j} \left \| \phi_j(\hat{y_t}) - \phi_j(y_t) \right \|^2_2 ,
\end{equation}
where $j$ refers to the output from layer $relu3\_3$ of the VGG-16 loss network, $\phi$.

\subsubsection{Style Loss}
The style representation of an image can be captured with the Gram matrix of CNN features\cite{perceptual_style_trans}. The Gram matrix at layer $j$ can be defined as

\begin{equation}
    \label{eqn:gram_matrix}
    G_j(x) = \phi^{\prime}_j(x)^T \phi^{\prime}_j (x) \in \mathbb{R} ^{C_j \times C_j},
\end{equation}
where $\phi^{\prime}_j(y) \in \mathbb{R}^{H_j W_j \times C_j}$ is the 2D matrix representation of the $j$th feature map obtained with a reshaping operation from $\phi_j(y)$.

Then, the style loss between the transformed image, $\hat{y_t}$, and the style image, $y_s$, is defined as
\begin{equation}
    \label{eqn:style_loss}
    \mathcal{L}_{\rm style}(\hat{y_t}, y_s) = \sum_{j\in S} \left \| \frac{G_j(\hat{x}) - G_j(y_s)}{H_j W_j C_j} \right \| ,
\end{equation}
where $S = \{relu1\_2, relu2\_2, relu3\_3, relu4\_3\}$ is the considered set of VGG-16 layers.

\subsection{Feature Learning}

Our feature-learning network, $f_\theta(\cdot)$, is adapted from the architecture presented by \cite{under_the_radar, mona_vision, self_sup_loc}, which is a U-Net style convolutional encoder-multi-decoder architecture to output keypoints, descriptors, and scores based on image inputs. The encoder is a VGG16 network~\cite{vgg16} pretrained on the ImageNet dataset~\cite{IMGNet}, truncated after the $conv\_5\_3$ layer. 
    
\begin{figure}[t]
    \centering
        \includegraphics[width=\linewidth]{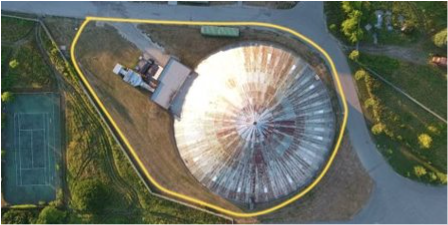}
        \caption{\textbf{UTIAS-In-the-Dark training paths.} The dataset contains 39 runs of a path collected at the campus of UTIAS in summer 2016. The same path was repeated hourly for 30 hours, which captures incremental lighting changes throughout the day.}

  \label{fig:training_testing_path}
\end{figure}

The keypoint-location decoder predicts the sub-pixel locations of each sparse 2D keypoint, $\textbf{q}=[u_l, v_l]^T$, in the left stereo image. To achieve this, we equally divide the image into $16 \times 16$ square cells, with each generating a single candidate keypoint. We then apply a spatial softmax on each cell and take a weighted average of the pixel coordinates to return the sub-pixel keypoint locations. 

In addition, the network computes the scores by applying a spatial softmax function to the output of the second decoder branch. The scores, $s$, are mapped to $[0,1]$, which predicts how useful a keypoint is for pose estimation. Finally, we generate dense descriptors for each pixel by resizing and concatenating the output of each encoder layer, which results in a descriptor vector, $\mathbf{d} \in \mathbb{R}^{960}$.

\subsubsection{Pose Estimation}
After generating $N$ keypoint predictions for the source image, we need to perform data association between the keypoints in the source and target images by performing a dense search for optimum keypoint locations in the target image. For each keypoint in the source image, we compute a matched point in the target image by taking the weighted sum of all image coordinates in the target image as follows:
\begin{equation}
    \label{eqn:kp_matching}
    \hat{\textbf{q}}^i_t = \sum^M_{j=1} \sigma(\tau f_{\rm zncc}(\textbf{d}_s^i, \textbf{d}_t^j))\textbf{q}_t^j,
\end{equation}
where M is the total number of pixels in the target image, $f_{\rm zncc}(\cdot)$ computes the zero-normalized cross correlation (ZNCC) between the descriptors, and $\sigma(\cdot)$ takes the temperature-weighted softmax with $\tau$ as the temperature. Finally, we find the descriptor, $\hat{\textbf{d}}_t^i$, and score, $\hat{s}^i_t$, for each computed target keypoint using bilinear interpolation. 

For the matched 2D keypoints, we compute their corresponding 3D coordinates using an inverse stereo camera model. The camera model, $\textbf{g}(\cdot)$, maps a 3D point, $\textbf{p}=[x\ y\ z]^T$, in the camera frame to a left stereo image coordinate, $\textbf{q}$, as follows:

\begin{equation}
    \label{eqn:stereo_model}
    \mbf{y} = \begin{bmatrix} u_l\\ v_l\\ d\end{bmatrix} = \begin{bmatrix}\textbf{q}\\ d
\end{bmatrix} = \textbf{g}(\textbf{p}) = \begin{bmatrix}
f_u & 0 & c_u &0 \\ 
0 & f_v & c_v & 0\\ 
0 & 0 & 0 & f_ub
\end{bmatrix}\frac{1}{z}\begin{bmatrix}
x\\ y\\ z\\ 1
\end{bmatrix},
\end{equation}
where $f_u$ and $f_v$ are the horizontal and vertical focal lengths in pixels, $c_u$ and $c_v$ are the camera's horizontal and vertical optical center coordinates in pixels, $d=u_l - u_r$ is the disparity obtained from stereo matching, and $b$ is the baseline in meters. We use the inverse stereo camera model to get each keypoint's 3D coordinates:
\begin{equation}
    \label{eqn:inverse_model}
    \textbf{p} = \begin{bmatrix}
x\\ 
y\\ 
z
\end{bmatrix} = \textbf{g}^{-1}(\mbf{y})=\frac{b}{d}\begin{bmatrix}
u_l-c_u\\ 
\frac{f_u}{f_v}(v_l-c_v)\\ 
f_u
\end{bmatrix}.
\end{equation}
Given the corresponding 2D keypoints $\{\textbf{q}_s^i, \hat{\textbf{q}}_t^i\}$ from the source and target images, we use (\ref{eqn:inverse_model}) to compute their 3D coordinates, $\{\textbf{p}_s^i, \hat{\textbf{p}}_t^i\}$. In addition, their descriptors and scores are $\{\textbf{d}_s^i, \hat{\textbf{d}}_t^i\}$ and $\{s_s^i, \hat{s}_t^i\}$.

We then use the ground-truth pose information to reject outliers, resulting in features that are geometrically consistent with each other. 

Given the inlier 3D keypoints, we can estimate the relative pose from the source to the target by minimizing the following cost using Singular Value Decomposition (SVD):

\begin{equation}
J = \sum^N_{i=1} w^i  \left \| (\textbf{C}_{ts}\textbf{p}^i_s + \textbf{r}^{st}_t) - \hat{\textbf{p}}^i_t  \right \|^2_2,
\end{equation}
where the weight for a matched point pair is a combination of the learned point scores and how well the descriptors match:
\begin{equation}
    w^i = \frac{1}{2} (f_{\rm zncc}(\textbf{d}_s^i, \hat{\textbf{d}}^i_t) + 1)s^i_s \hat{s}^i_t .
\end{equation}
Finally, we obtain the estimated relative transform between the source and target images as
\begin{equation}
\mathbf{\hat T}_{ts} = 
    \begin{bmatrix}
    \mathbf{\hat C}_{ts} & \mathbf{\hat r}^{st}_t \\ 
     \mathbf{0}^T & 1 
    \end{bmatrix}.
\end{equation}

\subsubsection{Keypoint Loss}

We transform the predicted source keypoints, $\textbf{p}_s^i$, to the target frame using the ground-truth relative transform, $\textbf{T}_{ts}$, and define a keypoint loss as follows:

\begin{equation}
    \label{eqn:keypoint_loss}
    \mathcal{L}_{\rm keypoint} = \sum^N_{i=1}{||\mathbf{T}_{ts} \mathbf{p}_s^i - \hat{\textbf{p}}_t^i ||^2_2}.
\end{equation}

\subsubsection{Pose Loss}
Given the ground-truth pose and the estimated pose, we can define pose loss as a weighted sum of the translational and rotational errors:
\begin{equation}
    \label{eqn:pose_loss}
    \mathcal{L}_{\rm pose} = \left \| \mathbf{r}^{st}_t - \hat{\textbf{r}}^{st}_t \right \|^2_2 + \lambda \left \| \mathbf{C}_{ts}  \mathbf{\hat C}_{ts}^T - \mathbf{1} \right \|^2_2, 
\end{equation}
where $\lambda$ is used to balance the rotational and translational errors.

The total loss is the weighted sum of all four losses (i.e., content loss, style loss, pose loss, and keypoint loss), which is defined as:
\begin{equation}
    \label{eqn:total_loss}
    \mathcal{L} = \lambda_1 \mathcal{L}_{\rm style}  + \lambda_2 \mathcal{L}_{\rm content} + \lambda_3 \mathcal{L}_{\rm pose} + \lambda_4 \mathcal{L}_{\rm keypoint},
\end{equation}
where the weights are determined empirically to balance the influence of the four loss terms.

\section{Experiments}
\begin{figure*}[h]
    \centering
        \includegraphics[width=1.0\textwidth]{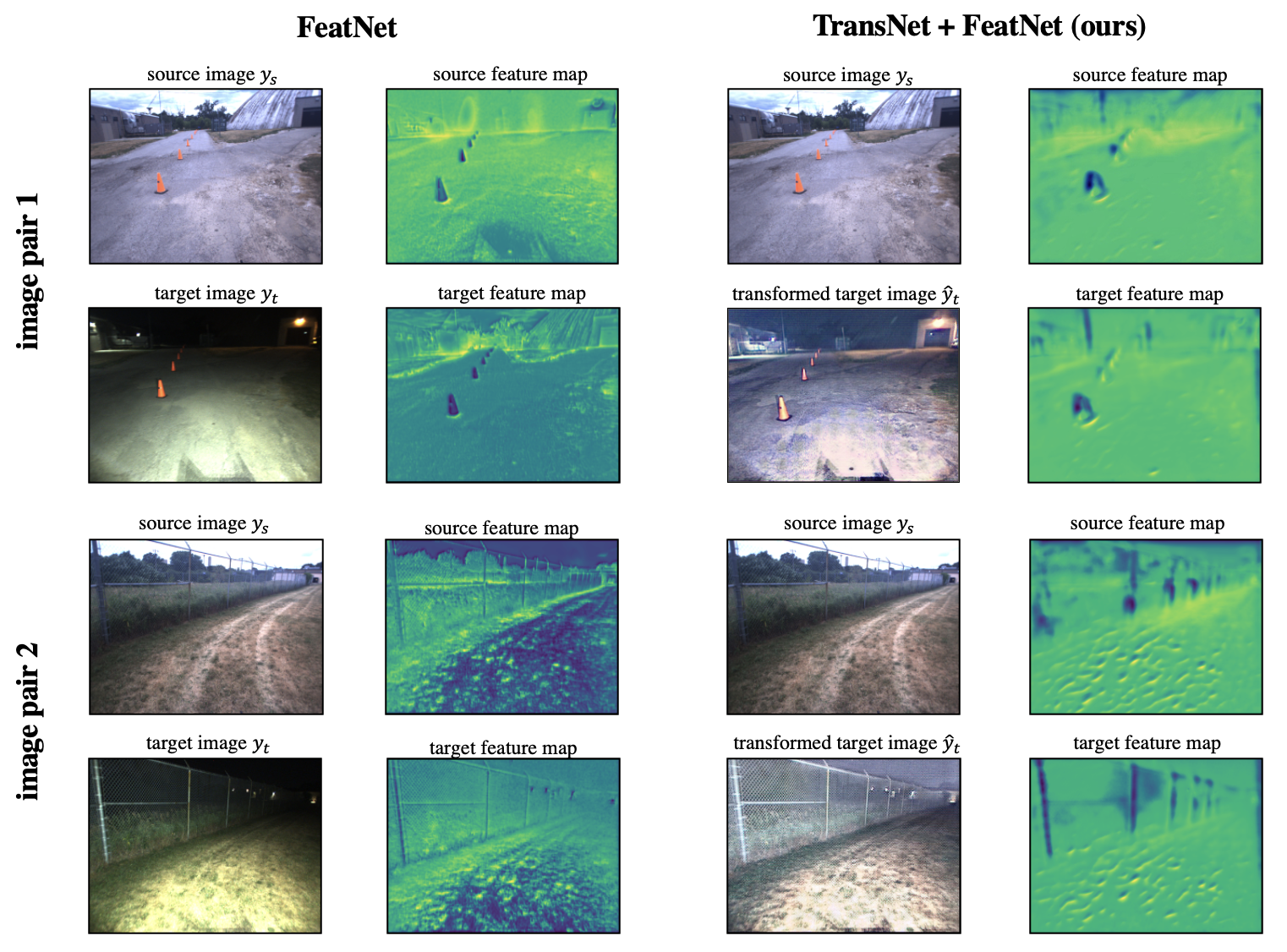}

    \caption{\textbf{Qualitative Results.} Visualization of the original source and target image pairs, the transformed target images and their feature maps. The feature maps are the dense feature detector values output from the FeatNet decoder, prior to the spatial softmax layer. }
    \label{fig:qualitative_results}
\end{figure*}

We conducted various experiments on real-world long-term vision datasets to validate and compare the effectiveness of feature learning and style transfer with selected baselines on localization tasks. We organize the experiments into the following three main categories for comprehensive analysis. In Section~\ref{sec:featnet_transNet}, we conducted four experiments on different combinations of FeatNet and TransNet with different training schemes to reveal the importance of our proposed joint training approach. In Section~\ref{sec:featnet_surf}, we compare the learned features against the classical SURF\cite{SURF} features to evaluate the effectiveness of our proposed approach in overcoming the limitations of traditional hand-crafted features. In Section~\ref{sec:transnet_colourspace}, we compare TransNet with the colourspace transformation proposed in~\cite{Clement_2020}. It aims to show the benefits of image transformation network brought to feature learning over the existing works, specifically in handling appearance changed caused by illumination variations.

\subsection{Datasets}
\vspace{-0.1cm}

For training, we use the public dataset, UTIAS-In-the-Dark, as described in~\cite{mona_vision}. The training data was collected using a Clearpath Grizzly robot with a Bumblebee XB3 camera. The robot autonomously repeats a path across drastic lighting changes using Multi-Experience VT\&R~\cite{MEL}. VT\&R stores stereo image keyframes as vertices in a spatio-temporal pose graph, where each edge contains the relative pose between a repeat vertex and the teach vertex. 
The dataset contains 39 runs of a path collected at the campus of UTIAS in summer 2016. The same path was repeated hourly for 30 hours, which captures incremental lighting changes throughout the day and night. The path for the In-The-Dark datasets can be visualized in Figure \ref{fig:training_testing_path}. To test for day-night image matching, we select one run that is collected during daytime as the reference sequence, and 9 runs in nighttime as the query sequences. We sample 20,000 image pairs for training, and 4,000 non-overlapping image pairs for testing. Each image pair consists of a daytime image as the source image and a nighttime image as the target image.

\subsection{Training and Inference}
We use the sampled image pairs to train the pipeline. For each input image pair, we use the daytime image as the style image to train the TransNet. For feature learning, we discard outliers based on keypoint error using the ground-truth pose during training and using RANSAC during inference. The FeatNet encoder is a VGG16 network \cite{vgg16} pretrained on the ImageNet dataset \cite{IMGNet}, truncated after the $conv\_5\_3$ layer. 
 Both networks are trained end-to-end on UTIAS-In-the-Dark using the Adam optimizer with a learning rate of $10^{-5}$ on an NVIDIA Tesla V100 DGXS GPU for 100 epochs. For our experiments, we set $\lambda_1=10^{-5}$, $\lambda_2=10^{-5}$, $\lambda_3=10.0$, $\lambda_4=2.0$.

\subsection{Comparison between TransNet and FeatNet }
\label{sec:featnet_transNet}

In this section, we evaluate different combinations of TransNet and FeatNet for metric localization under adverse conditions.  Specifically, we compare the performance of \textbf{\textit{FeatNet}}, \textbf{\textit{TransNet}}, \textbf{\textit{TransNet + FeatNet}}, and \textbf{$\textit{TransNet} \leftarrow \textit{FeatNet}$}. \textit{FeatNet} is the same approach as mentioned in ~\cite{mona_vision}, and is trained on the original day-night image pairs without performing style transfer. \textit{TransNet} is trained with only the perceptual losses proposed in~\cite{perceptual_style_trans}, and is tested on image pairs after integrating with a pretrained feature network to compute the localization errors. 
\textit{TransNet + FeatNet} is our proposed method, where both the TransNet and FeatNet are trained end-to-end on the feature losses and perceptual losses. Lastly, $\textit{TransNet} \leftarrow \textit{FeatNet}$ is similar to the previous method, but follows a two-stage training scheme. During the first stage, FeatNet is trained without image transformation. During the second stage, TransNet is trained with the FeatNet from the first stage, where the FeatNet stays fixed during the second-stage training.

For evaluation, we compute the path-following errors between the image pairs sampled from the UTIAS-In-the-Dark test set, then report the longitudinal errors, lateral errors and yaw angle errors in Table~\ref{tbl:comparison}.

When trained on the perceptual losses only, the transformed images from \textit{TransNet} are not optimized for keypoint matching, which is detrimental for visual metric localization as shown in Table~\ref{tbl:comparison}: the performance is significantly worse than when not doing style transfer at all (\textit{FeatNet}). 
When incorporating both of TransNet and FeatNet during training, TransNet is optimized explicitly for feature extraction, which leads to lower localization errors. However, we notice that the order of training both networks has little impact on the performance. Jointly training both networks (i.e., \textit{TransNet + FeatNet}) achieves comparable results as training FeatNet and TransNet sequentially (i.e.,  $\textit{TransNet} \leftarrow \textit{FeatNet}$). 



We visualize the transformed target images in Figure~\ref{fig:qualitative_results}, where the nighttime images are transformed into day-like conditions, making it easier for image matching between the day-night image pairs. In addition, we visualize the feature maps output from the FeatNet decoder prior to the spatial softmax layer, which can be interpreted as the detector response map for FeatNet. Adding image transformation results in more similar feature maps compared to using the FeatNet only, which improves the feature matching quality.

\subsection{Comparison between FeatNet and SURF}
\label{sec:featnet_surf}

In this section, we compare the effectiveness of using learned features against using classical SURF features~\cite{SURF}. Inspired by~\cite{adver_train_adver_cond}, we also integrate a differentiable SURF feature detector and descriptor pipeline with the TransNet. In addition to the style and content loss, we compute the L1 loss between SURF detector response maps and dense descriptor response maps on the transformed image pairs as described in~\cite{adver_train_adver_cond}.

From Table~\ref{tbl:comparison}, we can see that using an image transformation network improves the localization results from directly extracting SURF features on the original image pairs. However, using learned features significantly outperforms SURF features.

\subsection{Comparison between TransNet and Colourspace Transformation}
\label{sec:transnet_colourspace}

We compare the TransNet with other existing colourspace transformation networks. Clement et al.~\cite{Clement_2020} proposed to find an optimal colourspace transformation that performs nonlinear mapping from RGB to grayscale colourspaces to maximize the number of feature matches for image pairs. We compared our approach to two different formulations of colourspace transformations reported in \cite{Clement_2020}. Following the same naming convention, we refer to the generalized colour-constancy model as \textit{SumLog} and \textit{SumLog-E}, and refer to the MLP-based model as \textit{MLP-E}.

Since the detectors and matchers used in~\cite{Clement_2020} rely on non-differentiable components, a differentiable proxy network was used to predict the number of feature matches. For fair comparison, we replace their proxy matcher network by our pretrained FeatNet, which directly computes the pose and keypoint losses, to train the colourspace transformation network. 

From Table \ref{tbl:comparison}, we can see that the \textit{MLP-E} transformation outperforms the \textit{SumLog} and \textit{SumLog-E} transformations. However, using neural style transfer achieves consistently better results.

\begin{table}[t]

  \caption{\textbf{Comparison of the path-following errors and average number of inliers}. \textit{FeatNet} is pretrained without the image transformation component, whereas the proposed methods \textit{TransNet + FeatNet} are trained jointly and $\textit{TransNet} \leftarrow \textit{FeatNet}$ trains the FeatNet and TransNet sequentially. \textit{TransNet} and \textit{SumLog} are integrated with a pretrained FeatNet during inference time to compute the localization errors. 
  We report the longitudinal errors $\Delta x$ in meters, lateral errors $\Delta y$ in meters, the yaw angle errors $\Delta \theta$ in degrees, and the average number of feature inliers.} 
  \label{tbl:comparison}
  \centering
  \begin{tabu}{lcccc}
    \cmidrule[\heavyrulewidth]{2-5}
    &$\Delta x(m)$   &$\Delta y(m)$ & $ \Delta\theta(\degree)$ & inliers \\
    \cmidrule(r){1-5}
    TransNet~\cite{perceptual_style_trans} & 1.52 & 1.38 & 2.9 & 364 \\
    FeatNet\cite{mona_vision} & 0.07 & 0.05 & 0.47 & 484\\
    
    \textbf{TransNet \textleftarrow FeatNet} (ours) & \textbf{0.019} & \textbf{0.014} &0.25 & 504 \\ 
    \textbf{TransNet + FeatNet}  (ours) &0.024 & \textbf{0.014}&  \textbf{0.24} & \textbf{505} \\
    \midrule
        SURF\cite{SURF} & 1.24 & 1.52 & 2.84 & 40 \\
        TransNet+SURF\cite{adver_train_adver_cond} & 1.09& 1.20 & 1.82 & 59 \\
    \midrule
    SumLog\cite{Clement_2020} & 0.14 & 0.16 &1.67 & 463\\
    SumLog-E\textsuperscript{\textdagger}\cite{Clement_2020} & 0.12 & 0.26 & 1.02 & 474 \\
    MLP-E\textsuperscript{\textdagger}\cite{Clement_2020} & 0.08 & 0.12  &0.57 &490 \\
    
    \bottomrule
  \end{tabu}
\end{table}

\section{Conclusion}

In conclusion, we proposed a fully differentiable pipeline that consists of image transformation and feature learning, which improves the long-term metric localization performance under adverse appearance change. We performed various ablation studies to show the effectiveness of combining image transformation and feature learning in metric localization on real-world dataset. We discovered that the combination of neural style transfer and feature learning leads to the best results, which outperforms each individual component. Adopting neural style transfer prior to feature matching generates higher-quality matches, which subsequently leads to lower localization errors. 
For future extensions, we plan to integrate the proposed method into an existing VT\&R framework to test for real-time closed-loop deployment in unseen environments.

 \section*{ACKNOWLEDGMENT}
This work was supported by the Natural Sciences and Engineering Council (NSERC) of Canada.






{
\bibliographystyle{IEEEtran}
\bibliography{egbib}
}


\end{document}